%% file: main.tex
\documentclass{article}

\usepackage[final]{corl_2024} 
\usepackage{color}
\usepackage{lipsum}
\usepackage{graphicx}
\usepackage{booktabs}
\usepackage{amsmath} 
\usepackage{wrapfig}
\usepackage{algorithm}
\usepackage{algpseudocode}
\usepackage{amsmath}
\usepackage{amsfonts}
\usepackage{multirow}
\usepackage{enumitem}
\usepackage{subcaption}  
\usepackage{caption}
\input{macros}

\title{RoboCopilot: Human-in-the-loop Interactive Imitation Learning for Robot Manipulation}

\author{
    Philipp Wu$^{*,\dagger}$ \And
    Yide Shentu$^{*,\dagger}$ \And
    Qiayuan Liao$^{\dagger}$ \And
    Ding Jin \And
    Menglong Guo \And 
    Koushil Sreenath$^{\dagger}$ \And
    Xingyu Lin$^{\dagger, \ddagger}$ \And
    Pieter Abbeel$^{\dagger, \ddagger}$ \And
    *Equal contribution,
    $\ddagger$ Equal advising\\
    ${\dagger}$University of California, Berkeley \qquad  
}

\begin{document}

\makeatletter
\let\@oldmaketitle\@maketitle%
\renewcommand{\@maketitle}{\@oldmaketitle%
    \centering
    \vspace*{-4.0ex}
    \includegraphics[width=1.0\textwidth]{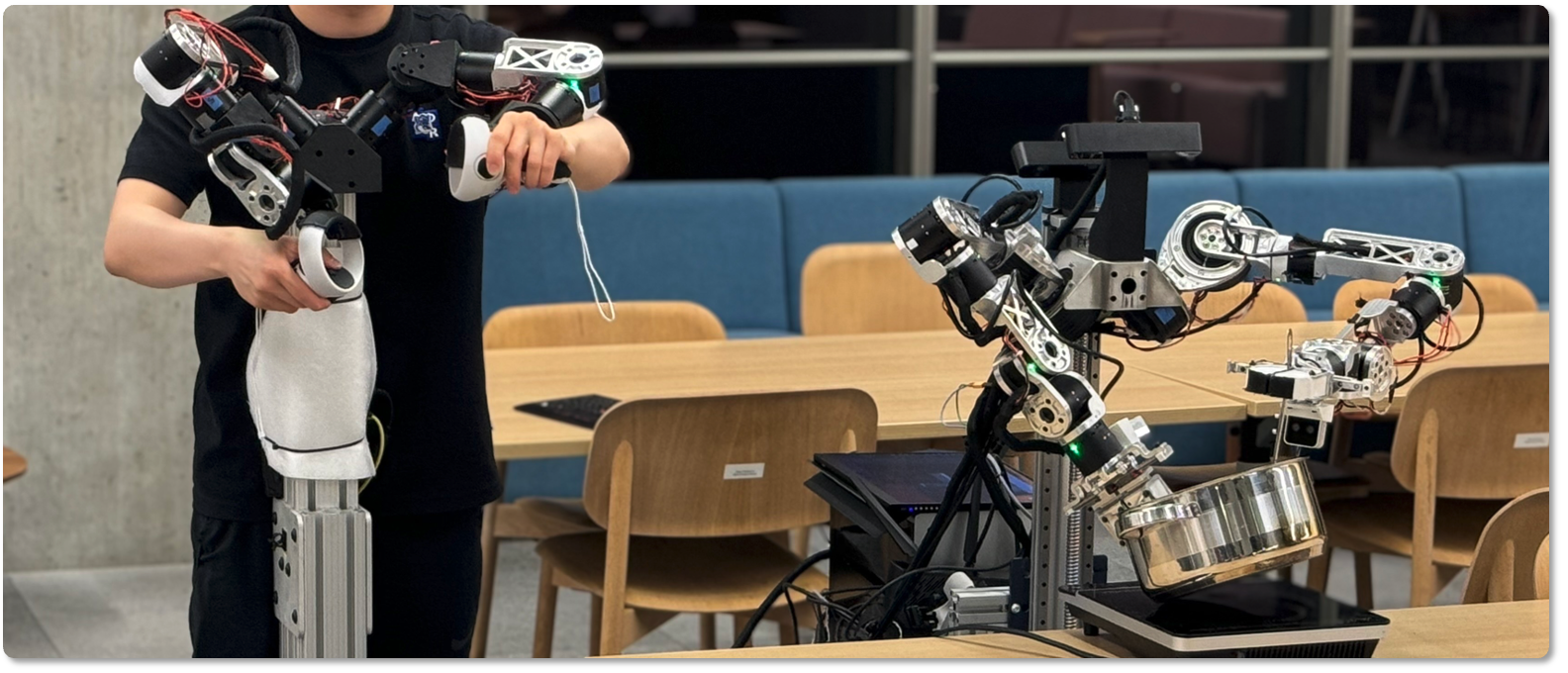}
    \captionof{figure}{A depiction of our RoboCopilot System which consists of a 20 degrees of freedom mobile bimanual robot and a bilateral teleoperation device. Our system enables easy teleoperation as well as human take-over at any time, allowing for an effective human-in-the-loop teleoperation system for interactive learning. }
    \label{fig:teaser}
    \vspace*{-0.5ex}
}
\makeatother

\maketitle


\begin{abstract}
Learning from human demonstration is an effective approach for learning complex manipulation skills. However, existing approaches heavily focus on learning from passive human demonstration data for its simplicity in data collection. Interactive human teaching has appealing theoretical and practical properties, but they are not well supported by existing human-robot interfaces. This paper proposes a novel system that enables seamless control switching between human and an autonomous policy for bi-manual manipulation tasks, enabling more efficient learning of new tasks. This is achieved through a compliant, bilateral teleoperation system. Through simulation and hardware experiments, we demonstrate the value of our system in an interactive human teaching for learning complex bi-manual manipulation skills. 
    
\end{abstract}

\keywords{Learning from Human Demonstration; Robot Manipulation} 

\section{Introduction}
\label{sec:introduction}
Humans play a critical role in teaching robot skills in robot manipulation. Recent developments in data-driven methods have shown promising results in learning complex robotic manipulation skills from human demonstration~\cite{zhao2023aloha, chi2023diffusion, wu2023gello, brohan2022rt1, jang2022bc, zhang2018deep}. The advantages of learning-from-demonstration approaches arise partly from their simplicity: humans provide a dataset of demonstration trajectories and then a policy is trained on the collected dataset to mimic the human actions. With enough human demonstrated data the policy will eventually learn to perform the same task like the human demonstrator. 

However, this passive imitation learning paradigm suffers from various inefficiencies that an interactive learning approach could overcome. In an interactive learning setting, humans take control of the robot when it fails and provide demonstrations for corrective behaviors. Prior works in Dataset Aggregation (DAgger) have shown, both theoretically and empirically, that corrective and interactive data improve policy learning performance by addressing the covariate shift issue~\cite{ross2011reduction,pomerleau1988alvinn, jang2021bc}. However, existing human teleoperation systems for manipulation are not designed for human intervention. Remote controllers such as spacemice and VR devices map the relative pose change to the robot end-effector, which is not intuitive for users, especially in an interactive setting where the human may need to take over at any time~\cite{zhang2018deep,wu2023gello}. Conversely, while exoskeleton-based devices have been effective in collecting passive human data~\cite{zhao2023aloha,wu2023gello,fang2023low}, synchronizing the policy rollout to an exoskeleton can be potentially unsafe for the human operator in a DAgger setting~\cite{kelly2019hgdagger}. Despite the recent advances of the easy-to-use teleoperation tools for multi-joints robot systems \cite{wu2023gello, zhao2023aloha}, the challenge of enabling intuitive interactive learning with these tools remains unresolved. 

In this paper, we introduce RoboCopilot, a robot system designed to accommodate \textbf{interactive human demonstrations} for learning bimanual robotic manipulation skills. We name our system RoboCopilot as it enables a human and a robot to cooperatively perform manipulation tasks, where the robot (the copilot) takes an assistive role before a lot of training. The system has two key components.
First, we adopt the algorithmic framework introduced by Human-Gated DAgger (HG-DAgger) for interactive imitation learning and adapt it to our manipulation setting. In this pipeline, we alternate between model training and data collection with a learned policy, where, during data collection, a human teleoperator interrupts and provides corrective feedback. This enables the robot to continually improve performance throughout the teaching process and collect higher-quality interactive data. We first verify the benefit of this procedure through controlled simulation experiments.
Second, we build custom hardware which consists of a mobile, compliant bimanual manipulator and a bilateral teleoperation device which enables us to instantiate our interactive learning pipeline. Our system enables a teleoperator to \textbf{seamlessly take over robot control} from a policy during the interactive training process. We show that with RoboCopilot, we can successfully teach the policy to learn long horizon, contact-rich, bimanual mobile manipulation skills.

\section{Related Work}
\label{sec:related_work}

\subsection{Imitation Learning from Passive Human Demonstrations}
Human demonstrations provide high-quality, action-labeled data for learning complex manipulation skills. Prior works show that behavior cloning is effective for learning from passive human data ~\cite{zhao2023aloha, chi2023diffusion, wu2023gello, brohan2022rt1,brohan2023rt2, haldar2023teach}, and increasing the number of trajectories consistently leads to better performance. However, it is challenging to provide performance guarantees for the trained policies. The number of demonstrations required is task-dependent, ranging from as few as ten demonstration trajectories for a simple picking skill~\cite{mandlekar2021matters} to up to a thousand trajectories~\cite{bousmalis2023robocat}. Furthermore, due to the issue of covariate shift, policies trained from passive offline data are unable to recover from the errors accumulated during online execution~\cite{ross2011reduction,spencer2021feedback}. Consequently, learning robust policies often has a significant hidden cost of an iterative process of data collection and policy learning. Our proposed system aims to enable interactive human teaching of complex manipulation skills, allowing the policy to continuously improve from newly collected demonstration data.

\input{figures/pipeline/figure}

\subsection{Interactive Imitation Learning}
In an interactive imitation learning setting, the student policy receives feedback from the expert policy during policy execution, enabling the student to continually improve and learn to correct rollout mistakes~\cite{celemin2022interactive}. Interactive learning approaches such as Data Aggregation (DAgger) have shown better sample efficiency and robustness compared to behavior cloning methods, both theoretically and in practice~\cite{ross2011reduction,pomerleau1988alvinn}. 
However, retroactively relabeling demonstrations, as originally formulated, can be quite difficult for a human, especially for robot manipulation \cite{grasping2016clutter, dart}.
Followup works have extended the DAgger framework to various settings to perform human-in-the-loop data collection~\cite{kelly2019hgdagger,hoque2021thriftydagger,li2022efficient,mandlekar2020human,liu2023robot,Spencer2020LearningFI,zhang2016queryefficientimitationlearningendtoend}. We follow Human-Gated DAgger (HG-DAgger)~\cite{kelly2019hgdagger} on its data collection phase, where the human operator determines when to intervene and take control of the student policy and when to release control back to the policy. However, such an approach requires the human and the agent to switch control when needed, which can be challenging for a typical teleoperation device. As such, the DAgger framework has not been widely adopted for complex bi-manual manipulation. RoboCopilot aims to bridge this gap through a teleoperation system that enables seamless human take over and continual policy improvement. 

\subsection{Robot Teleoperation Systems}
Teleoperation systems have existed from the very beginning of robotics~\cite{lichiardopol2007survey} and the design for the teleoperation interface heavily impacts the final performance for modern data driven methods. Controllers such as joysticks and VR controllers~\cite{zhang2018deep,zhu2022viola, dass2024telemoma,qin2023anyteleop}, while simple to set up, do not allow force feedback from the robot to the human operators and hide the kinematic joint limits of the robot from the human operators. Similarly, portable data collectors such as Universal Manipulation Interface  (UMI)~\cite{song2020grasping,young2020visual,chi2024universal,shafiullah2023bringing} also pose a gap by removing the kinematic constraints from the data collection process. While exoskeleton-based puppeteering solutions have shown impressive results in collecting and learning fine-grained bimanual manipulation policies~\cite{zhao2023aloha,wu2023gello,fu2024mobile,fang2023low}, the controllers lose synchronization during policy execution, making it difficult for human to intervene. Most related to ours, there has been a body of prior works on bilateral teleoperation~\cite{hulin2011dlr,katz_thesis,schwarz2021nimbro,elsner2022parti, DBLP:journals/corr/abs-2109-13382,FluidActuatorArm, TABLIS, Telexistence, bi1, bi2, bi3}. In this work, we build a compliant, bimanual teleoperation system and demonstrate its capability for interactive human teaching.

\section{Continual Interactive Imitation Learning}
\label{sec:method}


We aim to enable robots to continually learn from interactive demonstrations, where a human can take over control of the robot when the autonomous policy fails, as illustrated in Figure~\ref{fig:pipeline}. As human demonstration, robot execution and skill learning occur in a tight learning loop, this allows the human operator to understand where and when the autonomous policy fails, providing more targeted demonstrations on the model failure cases. We will summarize our interactive learning algorithm in this section and explain how we achieve this through our proposed RoboCopilot system in the next section.

\begin{wrapfigure}{R}{0.6\textwidth}
\vspace{-20pt}
   \begin{minipage}{0.6\textwidth}
    \begin{algorithm}[H]
    \small
    \caption{Human-in-the-loop Imitation Learning}
    \label{algo}
    \begin{algorithmic}[1]
    \State \textbf{Input:} Init policy $\pi_0$, human expert $\pi^*$, \# of iterations $N$, warmup human demonstration dataset $D$
    \State Train warmup policy $\pi_1$ on $D$ with behavior cloning 
    \For{$i = 2$ to $N$}
        \For{each step in the environment}
            \If{human expert $\pi^*$ intervenes}
                \State Execute expert human action: $\pi^*(s)$
                \State Add the intervention data: $D \leftarrow D \cup \{(s, \pi^*(s))\}$
            \Else
                \State Execute robot policy $\pi_i(s)$
            \EndIf
        \EndFor
        \State $\pi_{i+1}(s) \leftarrow finetune(\pi_{i}, D)$ 
    \EndFor
    \State \textbf{Output:} Final policy $\pi_N$
    \end{algorithmic}
    \end{algorithm}
  \end{minipage}
\vspace{-10pt}
\end{wrapfigure}

As shown in Algorithm~\ref{algo}, we first collect $K$ human demonstration trajectories to warm up the policy. Following HG-DAgger~\cite{mandlekar2020human}, we allow the human expert to determine when to intervene and take control of the robot execution. The intervention data from the human experts are added to the dataset. After each round of data collection, we fine-tune our policy using all the collected data by performing several gradient-based updates to the policy network. Our policy network is parameterized by a diffusion network~\cite{chi2023diffusion}. As the policy improves, the number of required interventions decreases significantly.

Prior work has shown that continually fine-tuning neural networks on data from non-stationary distributions causes catastrophic forgetting~\cite{goodfellow2013empirical, kirkpatrick2017overcoming}.
Despite this, we adopt fine-tuning during the interactive teaching process give its multiple advantages over training policies from scratch after the data collection is done: First, collecting online samples from the current policy alleviates covariate shift, as the human can intervene when necessary to teach correction skills. Second, as the policy is continually fine-tuned and executed, the human operator can quickly observe current failure modes and collect more targeted demonstrations. Finally, online learning reduces the overall training time. 

Practically, once the interactive teaching finishes, we can utilize the online interactive data and retraining a policy from scratch. Doing so avoids the non-stationarity of the training distribution but still takes advantage of the interactive data. In practice, we find this approach to provide the best policy that can be used for deployment. This amounts to restarting Algorithm~\ref{algo} with the most up to data dataset $D$.

\section{Teleoperation System}
\label{sec:hardware}

\begin{wrapfigure}{r}{0.65\textwidth}
\vspace{-10pt}
  \centering
  \includegraphics[width=0.65\textwidth]{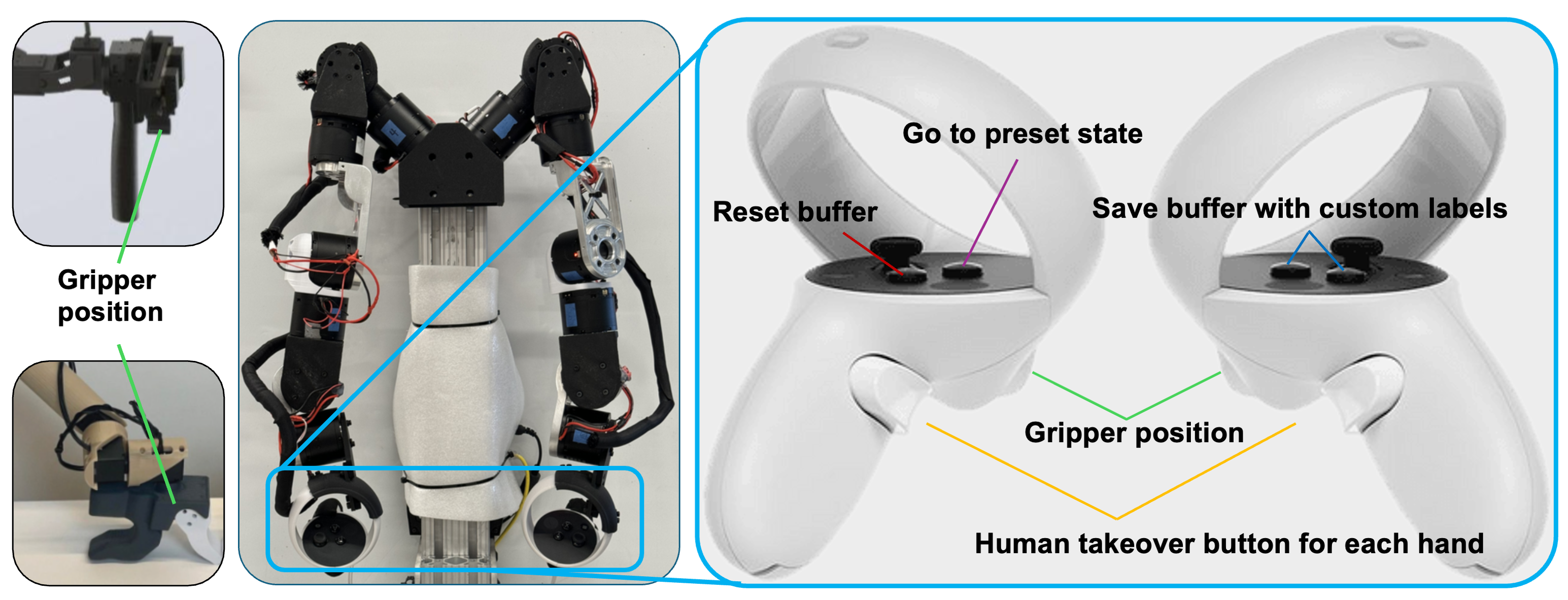}
  \caption{
    The comparison of the different end-effector human interface designs for different teleoperation systems is shown below: Left: Aloha \cite{zhao2023aloha} and GELLO \cite{wu2023gello}'s handheld interface. Middle: RoboCopilot layout, where we attached the Quest2 controller at the end of our GELLO device.
    Right: The key map of our end-effector human input interface. We optimized the layout to allow efficient gripper control and interactive human-in-the-loop teaching.
  }
  \label{fig:quest2}
\vspace{-10pt}
\end{wrapfigure}

We develop a versatile mobile bi-manual robot with a capable teleoperation system that allows us to instantiate human-in-the-loop interactive teaching for a wide range of real world tasks.
Our physical RoboCopilot teleoperation system consists of the physical robot, the teleoperation device, and the teleoperation workflow. More details about our custom low cost robot hardware can be found in Appendix \ref{sec:appendix_robot}.

Our teleoperation system is an effective way to enable human-in-the-loop imitation learning for manipulation.
A more capable and human intuitive device enables a user to more precisely control the robot which offers increased data quality for downstream learning.
While for dynamical systems with a lower dimensional action space like UAVs or cars, one can more easily apply DAgger algorithms by having a human relabel the data with a joystick 
 or wheel \cite{ross2012learningmonocularreactiveuav, kelly2019hgdagger}.
In general, there is no obvious way to relabel trajectories for manipulation trajectory data.
This is due in part to the high dimensional action space and complex environment dynamics.

\subsection{Teleoperation Device}
The core principle of the teleoperation leader device is to enable maximal data collection quality and ease of use for the human during operation and take over. Our puppeteering teleoperation device is a low cost approximate kinematic replica of our target arm, following GELLO \cite{wu2023gello}. More detailed hardware specifications are found in Appendix \ref{sec:appendix_robot}. We attach a Meta Quest 2 controller \cite{metaquest2} at the end of our teleoperation leader arms, serving as both an ergonomic gripping handle and a multi-input control device. Previous works have primarily focused on the active portion of controlling the arm, but our approach also considers the importance of a comprehensive and user-friendly interface. A comparison of different teleoperation devices is shown in Figure \ref{fig:quest2}. While teaching devices like ours are not new, the focus has largely been in an imitation learning context. Our goal is to show how to effectively extend the usefulness of these devices in an interactive learning setting.

\subsection{Active Control}
We actively control the motors of the leader arm enabling less user fatigue and smoother human take over. The weight and inertia of the teleoperation device requires additional energy input from the user to drive the device. Active gravity compensation using the manipulator equation alleviates this burden, allowing the operator to maneuver the system without excessive effort, reducing operator fatigue \cite{oussama07robotics,modernrobotics}. 

Active control also enables a bilateral system, where the leader arm can provide force feedback to the user felt by the follower arm. This enables a user to understand and feel the forces that the robot feels, allowing the user to more effectively operate the robot. We modify the PD control law used in \citet{katz_thesis} control law so the user feels a scaled down version of the forces 
 \cite{katz_thesis}. 
\vspace{-5mm}

\begin{align}
\tau_{L} &= \alpha K_p (\theta_{F} - \theta_{L}) + \beta K_d (\dot{\theta}_{F} - \dot{\theta}_{L}) \\
\tau_{F} &= K_p (\theta_{L} - \theta_{F}) + K_d (\dot{\theta}_{L} - \dot{\theta}_{F})
\end{align}

where $L$ is the leader teleoperation device and $F$ is the follower, $\tau$ is actuator torques, $\theta$ is joint position and $\dot{\theta}$  is joint velocity. We add $\alpha$ and $\beta$ constants so that the teleoperation device does not feel the full inertia of the arm, while still enabling effective control. Additionally, the use of active motors on our teleoperation device allows us to flexibly vary the $\alpha, \beta$ bilateral coefficient in different situations.

\subsection{Human-in-the-loop Interactive Learning}
Our RoboCopilot system is designed to be user-friendly and fully equipped with features that facilitate both the smooth operation of the robot for task completion and efficient data collection. The teleoperation device includes multiple buttons dedicated to data collection utilities, such as saving data and resetting the robot, streamlining the process of capturing valuable data during operation. This enables the teleoperator to be explicit about the data being collected, and can easily control for the data quality.

Control of the leader arm is intuitive, with bilateral control allowing the user to feel the forces experienced by the robot arm. This feature is particularly beneficial when the robot is in contact with the environment or handling heavy objects. When the autonomous policy is executing, the leader-follower relationship is reversed where the teleoperation device matches the real robot. The teleoperation can then activate take over at any point in time by engaging with the teleoperation device, without interrupting process flow as the devices are already synchronized. During robot policy execution, we use stiff gains on the leader to minimize the state difference, whereas during teleoperation mode, we use lower gains to ease the forces on the human teleoperator.

\section{Experiments}
\label{sec:results}
We conducted a series of experiments to demonstrate the value of our system for continual, interactive imitation learning in both simulation and the real world. 
Our system enables interactive data collection with a human in the loop. Our primary research question is how this interactive data collection affects the collected data quality. We measure the quality of the collected data by evaluating the final policy performance.
Specifically, we compare the policy performance using three data collection and policy learning methods: (1) Offline Behavior Cloning (BC), (2) Human-in-the-loop interactive learning where the policy is continually trained and used for collecting the next batch of data (Continual DAgger), and (3) Training a policy from scratch using all data, including the data collected by Continual DAgger (Batched Dagger). Through these experiments, we aim to demonstrate that interactive data collection can accelerate learning, reduce the need for laborious data collection process, and improve the overall performance. Additionally we demonstrate the effectiveness of our real world system at learning various tasks.
\input{tables/sim_appendix}
\subsection{Simulation}
We conduct imitation learning experiments on the standard Robomimic benchmark~\cite{mandlekar2021matters}. The benchmark provides human demonstration trajectories and simulation environments for a set of robotic manipulation tasks including bimanual tasks, showing the benefit of our approach for manipulation in general. The goal of these experiments is to validate the effectiveness of the algorithmic approach in a controlled reproducible environment. We choose the Can, Square, and Transport tasks from the benchmark which are visualized in Appendix \ref{sec:appendix_sim}. Experiments involving human interaction can be difficult to reproduce due to the volatile nature of human interaction. To improve reproducibility, we use the pre-trained expert diffusion policy $\pi^*$ from \citet{chi2023diffusion} as a human substitute. As this trained expert cannot decide when to intervene, we follow Algorithm \ref{algo} but use $\pi = (1-\beta)\pi_i + \beta\pi^*$ as our rollout policy when $|\pi_i(s_t) - \pi^*(s_t)| > \epsilon$. Here, $\pi_i$ is our training policy at the $i^{th}$ iteration.

The results are shown in Table~\ref{tab:sim_results}. Our DAgger variants use 10 expert demos to initialize, with the rest being DAgger-corrected trajectories.
Our continual learning method achieves superior performance to behavior cloning baselines, despite the issue of catastrophic forgetting. 
Some benefits of continual learning are hard to show with a learned policy trained on the distribution of human demos, as the policy does not have the ability to give sophisticated corrective demonstration nor intelligently collect targeted demonstrations in areas of failure. Even so, our simulation results demonstrate the value of interactive data, as seen by the performance boost provided by Continual DAgger.
This increase in data quality is further supported by the improvement of the policy in our Batched DAgger training where we train the policy from scratch using the Continual DAgger data. As expected, Batched DAgger greatly outperforms the continually trained variant in almost all cases. This greatly motivates the advantages of DAgger like training for manipulation. Next we test the effectiveness of our RoboCopilot system in real world settings using a human as the expert demonstrator, leveraging our teleoperation device to interrupt and interact with the scene.

\subsection{Real World}

Our real-world experiments provide similar conclusions to our simulation experiments, where interactive data collection provides higher quality data than passive data collection from the expert policy. We consistently observe that the success rate keeps increasing as we collect more interactive data, and the need for human intervention decreases accordingly. The need to intervene also serves as a useful signal in practice for when to stop collecting new data.
\input{tables/real_both}

\input{figures/full_task_descriptions/figure}
\subsubsection{Part Picking}
We test our RoboCopilot system in the real world on an industrial part picking task. The goal of the robot in this task is to pick up industrial aluminum extrusions and place them into a bin. We use two different sized aluminum extrusions, a small one which can be picked up with a single arm, a longer beam should be picked up by both arms. To control the experiment we define a domain of interest where the 8020 beams are placed. Additionally, for the placing bin, we mark 3 train locations, as well as 2 test locations. During testing of a policy, we choose and evaluate the robot of 18 predetermined locations. This task allows us to study various components of our system. First, the setting is very multimodal, as there may be different strategies and methods of how to pick up the object. Second we can test the effectiveness compliance of our system as holding the long beam with both hands is more straightforward for a compliant system. 

We define four ``task types": large bar, small bar left hand, small bar right hand, and small bar middle both hands. The primary question we aim to address is how effective our shared autonomy system is for collecting demonstrations to learn new skills. To start the experiment, we collect 12 trajectories (3 for each subtask) to initialize the policy, ensuring that it does not behave randomly or completely ineffectively. During the policy rollout, the human teleoperator has the option to take over from the autonomous system if the policy fails to complete the task. For each human-in-the-loop DAgger process, we will continue running the autonomous system until the human operator takes over, three times for each subtask. We tested our continuous learning pipeline under two different settings: Batched DAgger and Continual DAgger. For the Batched DAgger setting, every time we collect new interactive data, we relaunch the training process from scratch. In contrast, for the Continual DAgger setting, we continuously stream the human interactive data into the training pipeline to continuously fine-tune the model. All DAgger runs use policies from the warmup offline BC checkpoint. 

\subsubsection{Part Transport}
In addition to the stationary industrial part pick-and-place task, we applied the RoboCopilot pipeline to a more challenging industrial part transport task, which requires moving the robot base while performing manipulation tasks, as shown in Figure~\ref{fig:industrial_transport}. Specifically for this task, the robot must navigate to the tote, which is placed far from its starting position, to successfully complete the placing task.
See Appendix \ref{sec:appendix_real} for a visual depiction the task.
In Table~\ref{tab:real_results_8020}, we compare the performances of three methods given the same demonstration budget: Offline BC, which trains a policy from passively collected data; Continual DAgger, which continually fine-tunes a policy after collecting 10, 15, and 20 trajectories; and Batched DAgger, which uses the final dataset obtained by running Continual DAgger and trains a policy from scratch. During Continual DAgger, we observe that humans can identify initial configurations where the current policy fails and reset intelligently. In this way, we find that Continual DAgger is more data-efficient than Offline BC, thanks to intelligent reset. However, Continual DAgger keeps fine-tuning the same policy as new demonstrations are added to the dataset, leading to suboptimal policy optimization. In Batched DAgger, we train one policy from scratch using this interactively collected dataset and achieve the best performance, further validating our claim that human-in-the-loop data collection improves the quality of the collected data.
\input{figures/task/kitchen}
\subsubsection{Kitchen}

Lastly we study a long horizon task in a toy kitchen, requiring the robot to execute multiple individual steps in order to ``cook the tomato". This high level task is broken up into 4 smaller subtasks, where the robot first must open the cabinet door, then pick the tomato, then put the tomato in the pot, and lastly turn the dial. If the robot fails to accomplish any one task, we count failure for all the following steps as in the CALVIN long horizon benchmark \citep{mees2022calvinbenchmarklanguageconditionedpolicy}. See Fig \ref{fig:kitchen_task} for an illustration of this task. We compare naive Offline BC to our interactive approach leveraging Batched DAgger.
The results are found in Table \ref{tab:real_kitchen}.
We find that leveraging human-in-the-loop not only results in better long horizon performance, but it also decreases the necessary human teleoperated time to reach that performance. 
Interestingly we find that more demos for Offline BC does not help the policy learn to fully complete the task, whereas our Batched DAgger approach enables a significant success increase in fully completing the sequence of subtasks, despite having a smaller overall dataset size. 
Our experiments show that Batched DAgger with one iteration of data collection and a total of 30 trajectories is able to match or outperform the success rates of Offline BC with 45 trajectories on all subtasks, with the performance gap being the largest for completing the full task (i.e. Turn Dial). This shows that our approach is also effective in learning trajectories that have long time horizons.
Lastly we see that Batched DAgger requires less human collected trajectory steps for a fixed number of training trajectories, as the human only needs to operate after a policy failure. This reduction in human intervention steps further underscores the efficiency of our human-in-the-loop system in teaching the robot 


\input{tables/kitchen_real}


\section{Discussion}
\label{sec:discussion}

\textbf{Limitations.}
While our RoboCopilot system shows promising results in enabling interactive teaching and continual learning for bimanual robotic manipulation, it is not without limitations. First, the requirement for human intervention during the learning process can be resource-intensive, as it necessitates a skilled operator to oversee and correct the robot's actions continually. This dependency might limit the scalability of our system in environments where such expertise is not readily available or the task is too difficult to even teleoperate. Second, although our system aims to be cost-efficient, the initial setup and maintenance costs might still be prohibitive for smaller organizations or research labs due to the large number of custom components. Additionally, the use of planetary gearboxes, while beneficial for compliance and cost, introduces some degree of backlash, which may affect the precision of fine manipulation tasks.

\textbf{Conclusions}
In this work, we introduced RoboCopilot, a novel system that enables human-in-the-loop interactive imitation learning for bimanual robotic manipulation tasks. The RoboCopilot system underscores the potential of seamlessly interfacing a human operator and an autonomous policy for interactive teaching. Future work will focus on enhancing its scalability and ease of deployment in diverse real-world environments.


\clearpage
\acknowledgments{.Pieter Abbeel holds concurrent appointments as a Professor at UC Berkeley and as an Amazon Scholar. This paper describes work performed at UC Berkeley and is not associated with Amazon.}


\bibliography{references}  

\clearpage
\appendix
\input{appendix}

\end{document}

%% file: macros.tex
\definecolor{MyDarkRed}{rgb}{0.8,0.02,0.02}
\definecolor{MyDarkGreen}{rgb}{0.02,0.6,0.02}
\definecolor{MyPurple}{rgb}{0.6,0.1,.9}



%% file: figures/pipeline/figure.tex
\begin{figure}[ht]
\centering
\begin{subfigure}{\textwidth}
    \includegraphics[width=\textwidth]{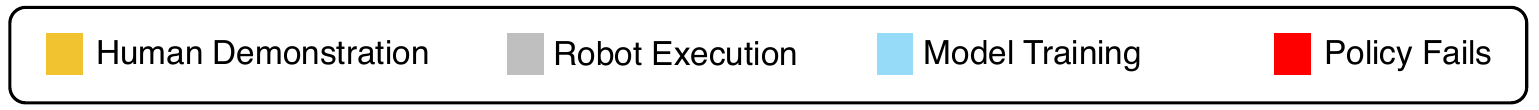}
\end{subfigure}
\hfill
\begin{subfigure}{0.49\textwidth}
    \includegraphics[width=\textwidth]{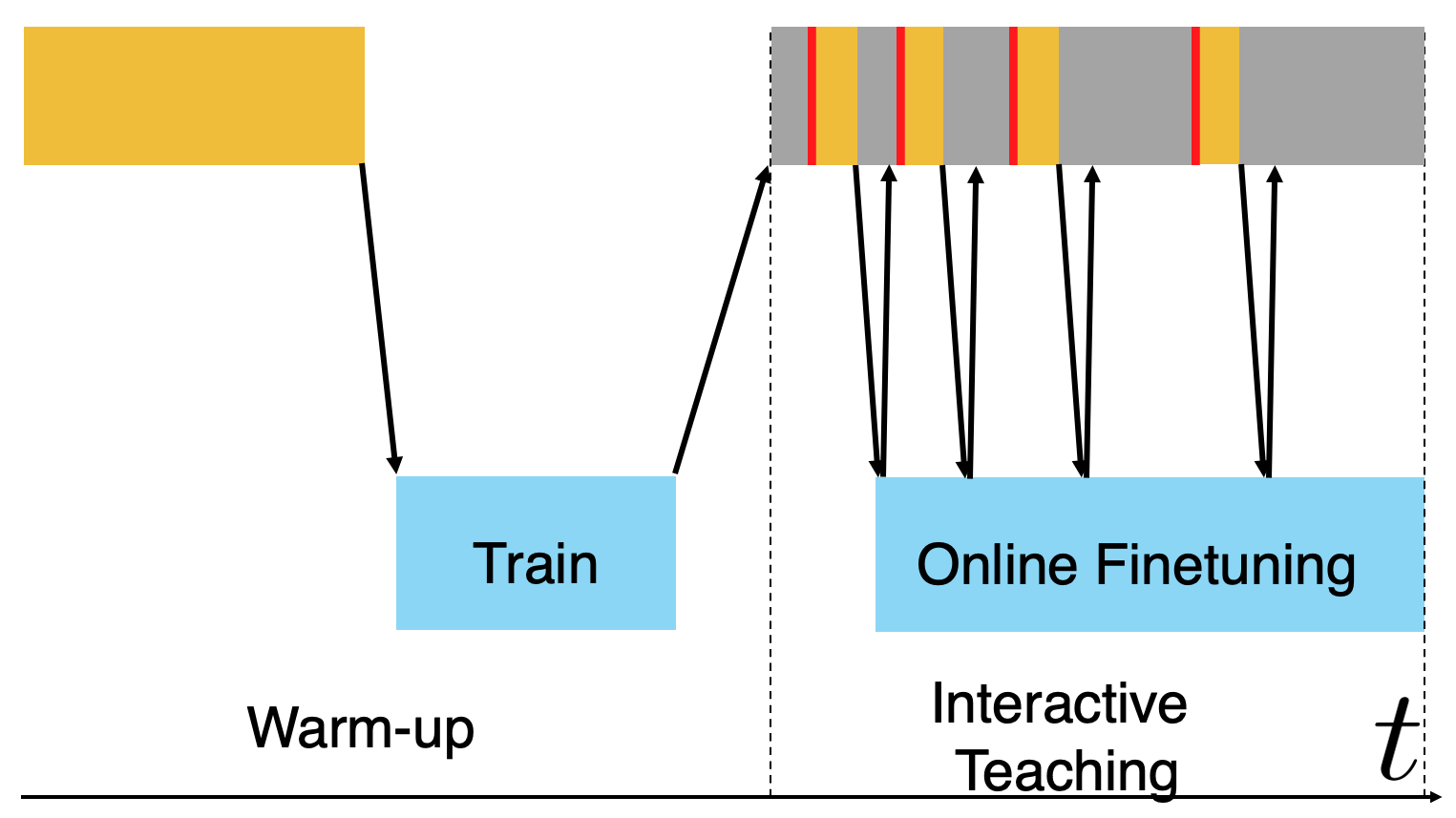}
    \caption{Workflow for learning one skill. }
    \label{fig:first}
\end{subfigure}
\hfill
\begin{subfigure}{0.49\textwidth}
    \includegraphics[width=\textwidth]{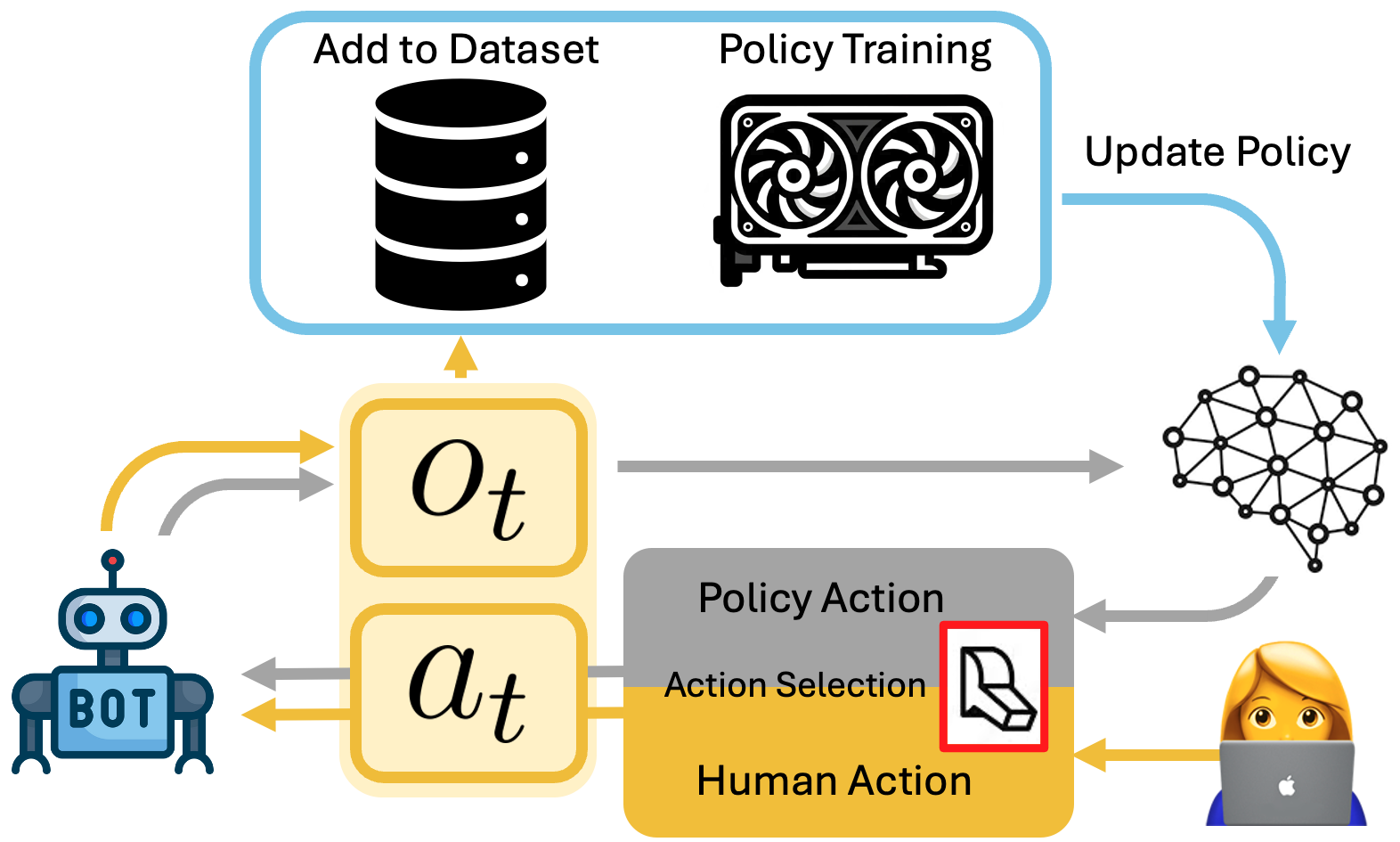}
    \caption{Dataflow during interactive teaching.}
    \label{fig:second}
\end{subfigure}
\hfill
\caption{Overview for our interactive teaching system. (a) Workflow for learning a single skill:  We start with a set of human demonstrations to pre-train the initial policy. Then in the interactive teaching stage, the policy is deployed and a human intervenes upon policy failure. The policy is continually fine-tuned from these new demos. As policy performance improves, less human intervention is needed. (b) During robot execution, the robot policy takes sensor observations and outputs actions. The human can decide when to use the policy switch to teleoperation. This enables the human to interrupt the robot on policy failure and correct the mistake, storing the data into the dataset. The model is continually training and being updated.}
\label{fig:pipeline}
\end{figure}

%% file: tables/sim_appendix.tex
\begin{table*}[t]
  \small
  \caption{
  Simulation results comparing behavior cloning and various DAgger methods on the robomimic benchmark \cite{mandlekar2021matters}. We collect 10 trajectories for warming up the dataset and training the initial policy. BC-H refers to behavior cloning from a human demonstration dataset. BC-P refers to behavior cloning from an expert dataset collected by our expect policy. The expert policy is the same policy that makes interventions in our DAgger methods. \textbf{Continual DAgger} denotes our continual DAgger method that continually finetuning with interaction data. \textbf{Batched DAgger} denotes training from scratch with the Continual DAgger collected data. We see leveraging human in the loop during the data collection process significantly improves over the baselines. For our DAgger runs, we initialize with 10 expert demo trajectories. Experiments are run across three seeds. 
}
  \label{tab:sim_results}
  \centering
  \begin{tabular}{l l l l l l}
    \toprule
\multirow{2}{*}{\shortstack{Environment}} & Total Number of & \multirow{2}{*}{\shortstack{BC-H}} &  \multirow{2}{*}{\shortstack{BC-P}} & \textbf{Continual} & \textbf{Batched} \\
            & Trajectories    &                                    &         & \textbf{DAgger (Ours)} & \textbf{DAgger (Ours)} \\
    \midrule
      Can & 10 (warmup)    & 0.28±0.04 &  0.43±0.0 &                --- &                --- \\
      Can & 15 & 0.45±0.02 & 0.57±0.04 &          0.57±0.02 & \textbf{0.63±0.02} \\
      Can & 20 & 0.61±0.04 & 0.62±0.02 &          0.62±0.02 & \textbf{0.72±0.02} \\
      Can & 30 & 0.76±0.01 & 0.82±0.02 & \textbf{0.85±0.04} &          0.83±0.02 \\
      Can & 40 & 0.83±0.03 & \textbf{0.93±0.01} &          0.87±0.02 & \textbf{0.93±0.02} \\
    \midrule
   Square & 10 (warmup)    & 0.26±0.02 & 0.30±0.02 &                --- &                --- \\
   Square & 15 & 0.38±0.05 & 0.28±0.03 &          0.39±0.02 &         \textbf{0.43±0.02} \\
   Square & 20 & 0.40±0.00 & 0.41±0.03 &          0.43±0.01 &         \textbf{0.44±0.03} \\
   Square & 30 & 0.48±0.05 & 0.46±0.03 &          0.47±0.02 &         \textbf{0.55±0.02} \\
   Square & 40 & 0.52±0.01 & 0.51±0.02 &          0.53±0.05 &          \textbf{0.57±0.01} \\
    \midrule
Transport & 10 (warmup)    & 0.41±0.10 & 0.65±0.04 &                --- &                --- \\
Transport & 15 & 0.55±0.02 & 0.57±0.05 &          0.72±0.06 &          \textbf{0.87±0.02} \\
Transport & 20 & 0.53±0.05 & 0.70±0.02 &          0.88±0.02 &          \textbf{0.97±0.02} \\
Transport & 30 & 0.65±0.04 & 0.61±0.08 &          0.90±0.04 &          \textbf{0.92±0.02} \\
Transport & 40 & 0.77±0.04 & 0.83±0.01 &          0.90±0.04 &          \textbf{0.93±0.02} \\
    \bottomrule

  \end{tabular}
\end{table*}

%% file: tables/real_both.tex
\begin{table*}
  \small
  \caption{Task completion rate of the industrial picking experiment we ran in the real environment. From this table, we can see that under the human-in-the-loop interactive learning setting, the policy can achieve higher performance given the same amount of human-labeled trajectories. With the human-in-the-loop copilot data
collection process, the learned policy achieves a much higher success rate given the same number of training
trajectories. Additionally, since the human operator can allow the policy to take over again once a mistake
is corrected, the number of human-teleoperated timesteps in these DAggered trajectories is significantly less
compared to those in offline BC data.
  }
  \label{tab:real_results_8020}
  \centering
  \begin{tabular}{c c c c c}
    \toprule
    Task                     & \# of Trajectories & Offline BC & Continual DAgger (Ours) & Batched DAgger (Ours)\\
    \midrule
    \multirow{3}{*}{\shortstack{Industrial\\Picking}} & 12 (warmup) & 38.9\% & --- & --- \\
                                        & 24 & 61.1\% & \textbf{72.2\%} & \textbf{72.2\%} \\
                                        & 36 & 66.7\% & 72.2\% & \textbf{77.7\%} \\
    \midrule
    \multirow{3}{*}{\shortstack{Industrial\\Part Transport}} & 10 (warmup) & 33.3\% & --- & --- \\
                   & 15 & 50.0\% & 66.7\% & \textbf{75.0\%} \\
                   & 20 & 58.3\% & 75.0\% & \textbf{91.7\%} \\
    \bottomrule
  \end{tabular}
  
\end{table*}

%% file: figures/full_task_descriptions/figure.tex
\begin{figure}[h]
  \centering
  \includegraphics[width=\textwidth]{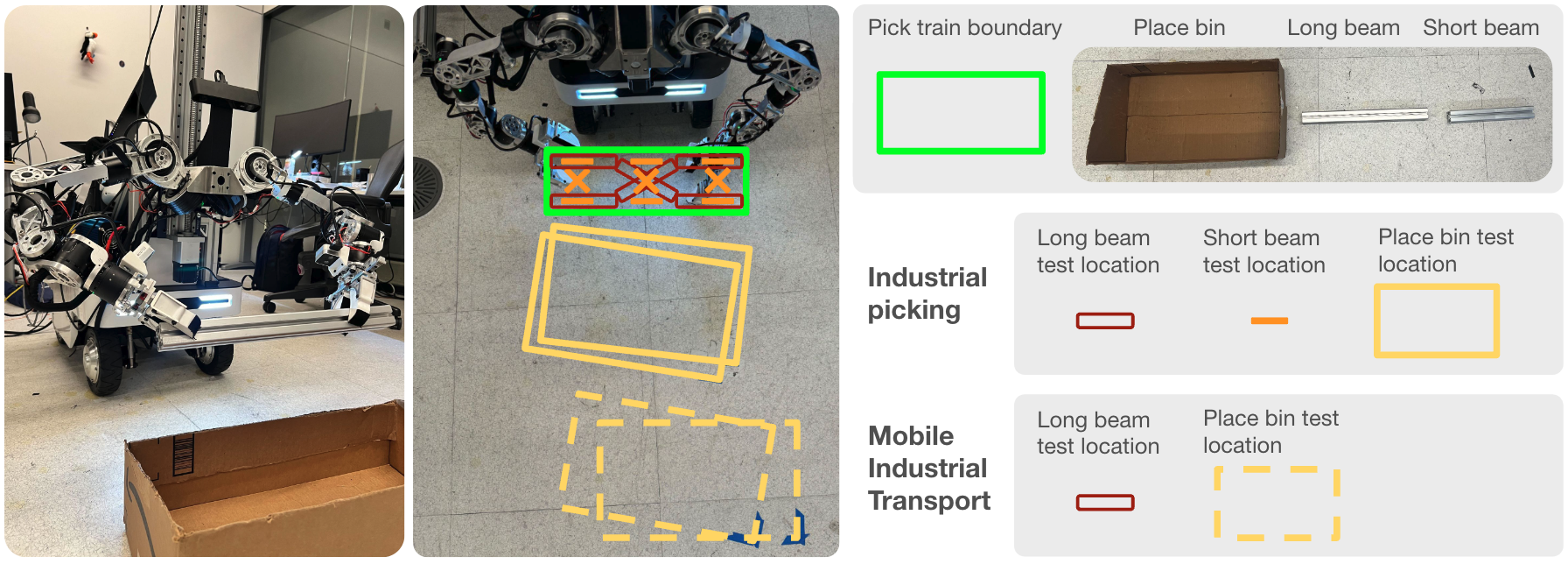}
  \caption{
  An illustration of the evaluation protocol for the industrial part transport tasks. During training, the beam is placed in different positions within a defined boundary (highlighted in green). Industrial picking requires the robot to locate and manipulate the long beam or short beam and place it within the bin. Mobile industrial picking only considers the long beam, but the bin is further away, requiring the robot to drive the base before placing. We label the poses of the beams and the bin to ensure consistency during evaluation.
  }
  \label{fig:full_task_description}
\end{figure}

%% file: figures/task/kitchen.tex
\begin{figure}[b]
  \centering
  \includegraphics[width=1.0\textwidth]{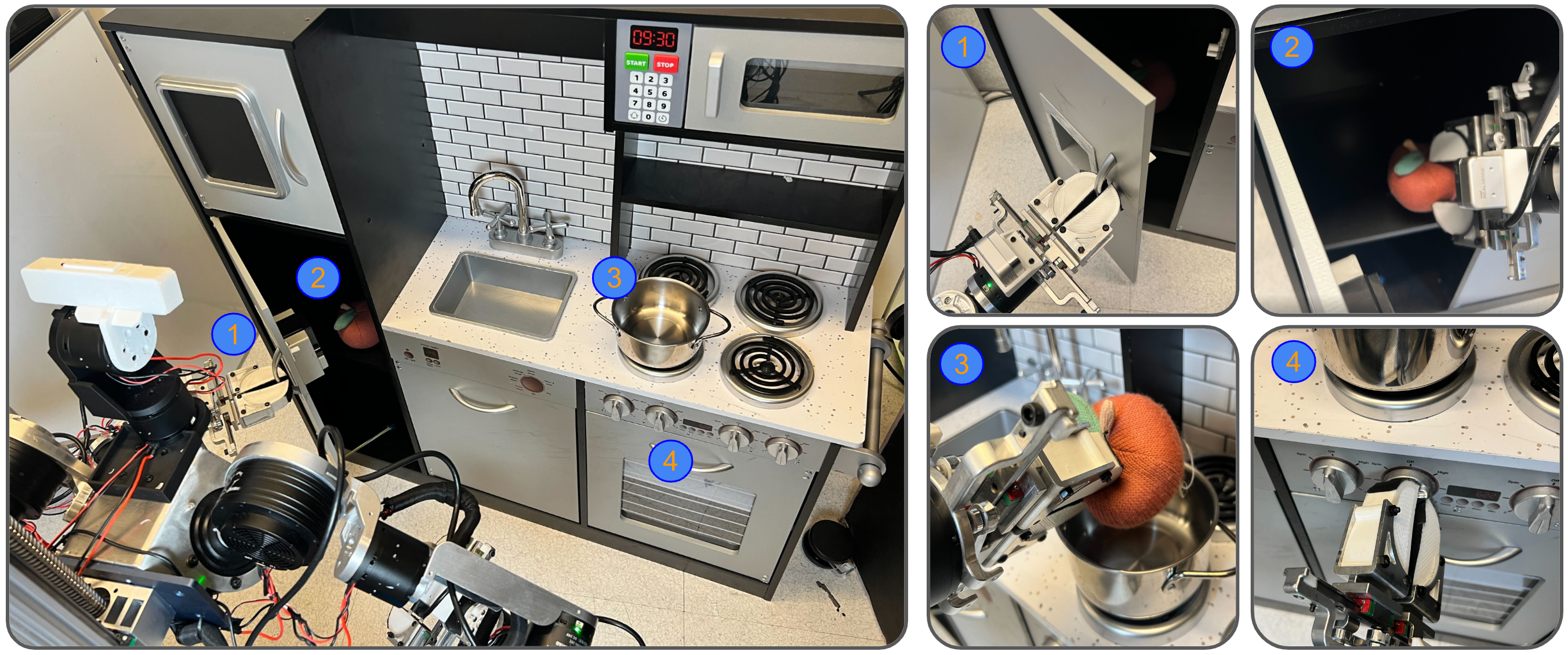}
  \caption{
  An overall illustration of the toy kitchen task. (Subtask 1) The robot needs to first open the spring-loaded cabinet door and hold the door open. (Subtask 2) The robot can pick the tomato and transfer it to the stove area. (Subtask 3): At this stage, the robot needs to put the tomato into the pot. Notice that the pot can be at different locations. (Subtask 4) Finally, the robot needs to turn the correct dial depending on which stove the pot is at.}
  \label{fig:kitchen_task}
\end{figure}

%% file: tables/kitchen_real.tex
\begin{table*}
  \small
  \caption{Experimental results on the long horizon kitchen task. We compare Offline BC with Batched DAgger for different trajectory dataset sizes, in increments of 15. The size of the training dataset is shown. For each experiment, we also show the number steps that required a human to operate the robot to collect the data at each stage. We also show a detailed breakdown of the task success rate for each subtask in the overall task.}
  \label{tab:real_kitchen}
  \centering
  \begin{tabular}{c l c c c c c c}
    \toprule
    \# of        & \multirow{2}{*}{\shortstack{Method}} & Dataset Size    & \% intervention & Open  & Grab   & Place  & Turn \\
    Trajectories &                                      & (\# Transitions)& steps & Door  & Tomato & Tomato & Dial \\
    \midrule                                                               
    15                               & Offline BC       &  8491           & 100\% & 100\% & 70\% & 25\%  & 5\% \\
    \midrule                                                                          
    \multirow{2}{*}{\shortstack{30}} & Offline BC       & 16654           & 100\% & 100\% & 70\% & 35\%  & 5\% \\
                                     & Batched DAgger   & 15162           & 65 \% & 100\% & 85\% & 45\%  & 20\% \\
    \midrule                                                                 
    \multirow{2}{*}{\shortstack{45}} &  Offline BC      &  25694          & 100\% & 100\% & 85\% & 40\% & 0\% \\
                                     &  Batched DAgger  &  20844          & 52\%  & 100\% & 90\% & 50\% & 30\% \\
    \bottomrule
  \end{tabular}
\end{table*}

%% file: appendix.tex
\section{Additional Real-world Task Visualization}
\label{sec:appendix_real}

\begin{figure}[h]
  \centering
  \includegraphics[width=0.84\textwidth]{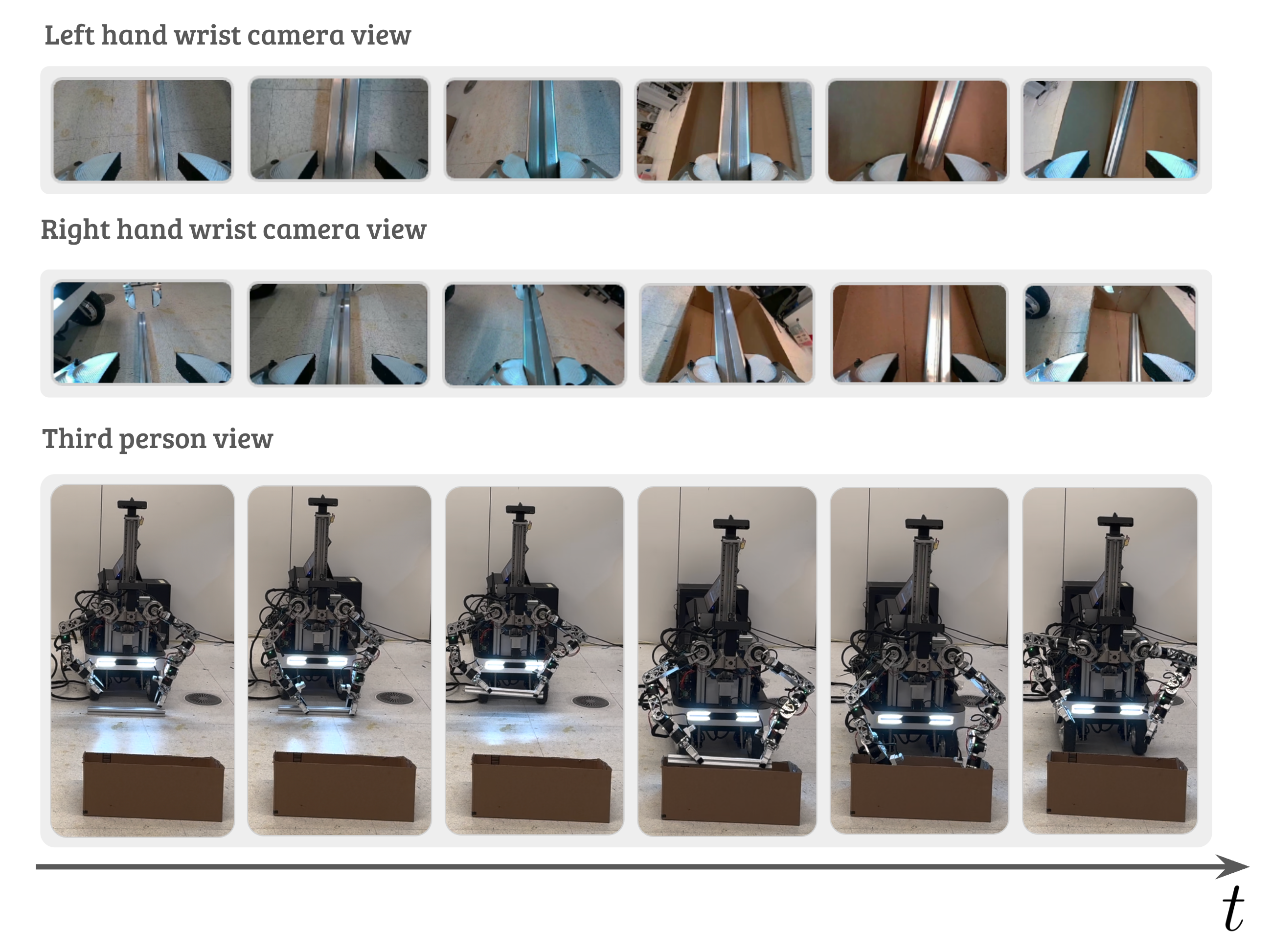}
  \caption{A step-by-step illustration of the industrial part transport task. The robot first needs to pick up the heavy industrial part from the floor using both hands. It then searches for and moves towards the tote. Finally, the robot must accurately drop the industrial part into the tote. The first two rows display the left and right wrist camera views along one autonomous trajectory. The bottom row shows a third-person view of the entire robot system during task execution.}
  \label{fig:industrial_transport}
\end{figure}


\section{Simulation Task Visualizations}
\label{sec:appendix_sim}
\input{figures/appendix_sim_tasks/figure}

\newpage

\section{Teleoperation Capabilities}
\label{sec:appendix_capabilities}
Our proposed RoboCopilot system not only enables efficient interactive teaching but also enhances teleoperation capabilities beyond the unilateral teleoperation devices \cite{zhao2023aloha,wu2023gello}. This improvement is achieved through active joint motor selection, which provides inertia compensation and bilateral feedback. Additionally, our fully compliant arm design allows for contact-rich bimanual tasks without compromising safety or risking damage to the robot hardware. In \autoref{fig:teleoperated_tasks} below, we demonstrate several daily kitchen tasks performed via teleoperation to showcase our system’s capabilities.
\begin{figure}[h]
  \centering
  \includegraphics[width=0.99\textwidth]{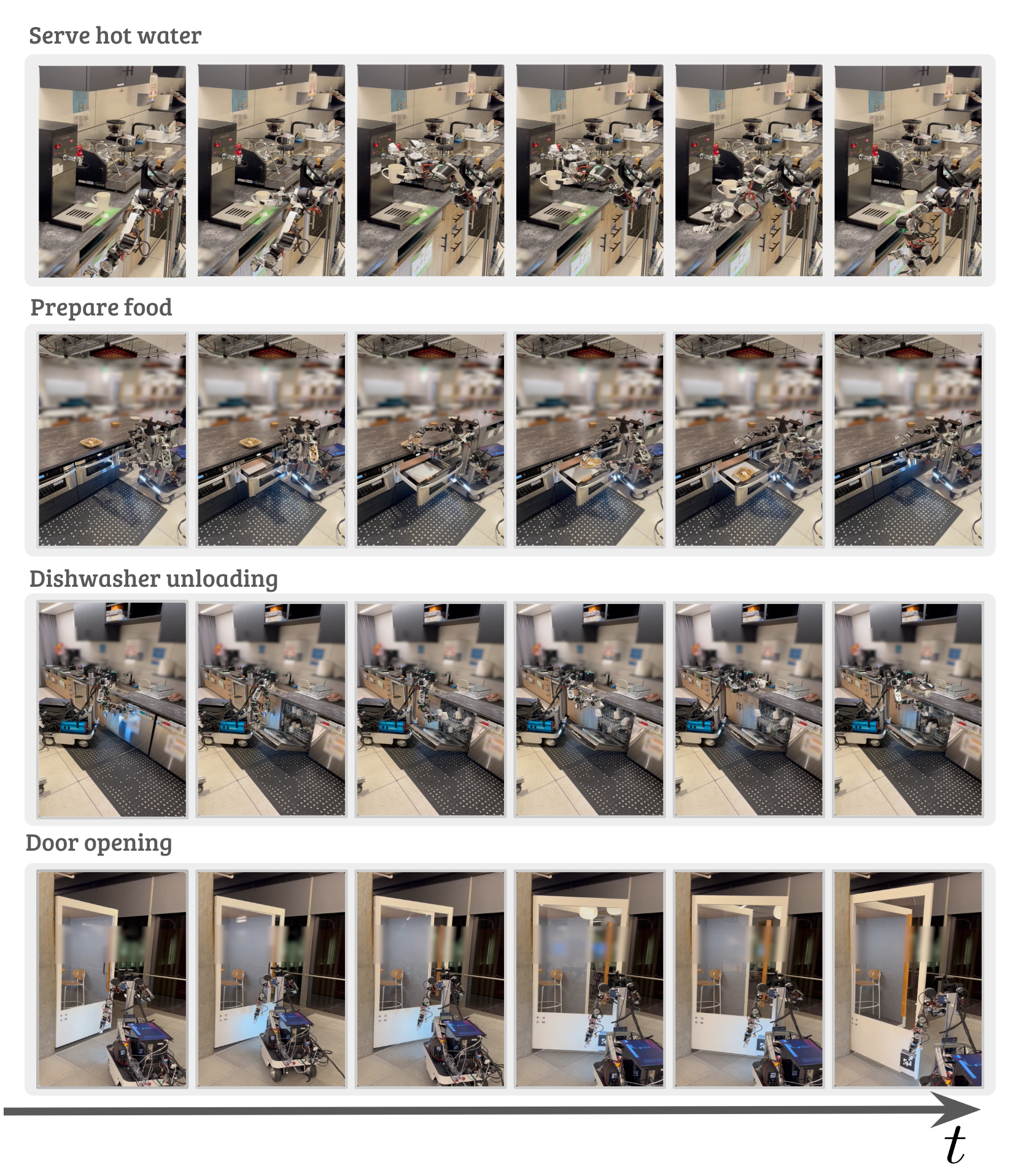}
  \caption{Teleoperated Execution of Various Contact-Rich and Whole-Body Control Tasks. The RoboCopilot system demonstrates its proficiency in performing a range of daily kitchen tasks under human teleoperation. These tasks include serving hot water, preparing food, unloading the dishwasher, and opening doors. The robot's fully compliant design enables safe and effective handling of contact-rich interactions, showcasing its ability to perform a wide range of complicated tasks.}
  \label{fig:teleoperated_tasks}
\end{figure}

\newpage

\section{Robot Hardware Details}
\label{sec:appendix_robot}

The design space for such a complex system is large. Thus we utilize the following three key principles the development for our RoboCopliot system:
\begin{itemize}[left=0pt]
    \item \textbf{Safe:} The robot hardware must be safe for humans, the environment, and itself. Compliance, a mechanical property that describes the ability of the robot to respond to external forces, is critical to achieving this safety.
    \item \textbf{Easy to Use:} To facilitate efficient interactive learning, our system needs a capable teleoperation system that allows human operators to perform complex and contact-rich tasks. Additionally, this system must enable the human operator to take control of the robot seamlessly when necessary. 
    \item \textbf{Accessible:} Our design aims to be cost-efficient and simple while still being able to perform a wide range of daily tasks. We believe accessibility is crucial for scaling up data collection and robot deployment in real-world scenarios.
\end{itemize}

Our robot, guided by the key principles above, is a custom low-cost mobile bimanual manipulator designed for everyday tasks. It consists of 7 DOF arms with parallel jaw grippers, an actuatable torso, and an omnidirectional base, totaling 20 degrees of freedom. We use two realsense D405 cameras \cite{intelrealsense} mounted on the wrist which are used to provide camera observations to the robot.

\textbf{Compliance from Quasi-Direct Drive Actuators.}
Compliance allows the robot to feel when it is perturbed by an external force, which is crucial for the robot to safely work with humans and perform complicated tasks in environments where there is high uncertainty. Compliance can be achieved with the addition of force torque sensors by measuring and responding to external forces in a tight control loop \cite{series_elastic, DLR_robot}. Compliance can also be achieved passively through a backdrivable transmission, where the system can be easily driven by external force without active control \cite{Wyrobek2008TowardsAP, gealy2019quasidirect, quigley}.
Taking cost and simplicity into consideration we achieve compliance with off-the-shelf mini-cheetah style \cite{Katz2018ALC} quasi direct drive (QDD) transmissions which achieve backdrivability through a low gear ratio planetary gearbox.
In this setup, external forces applied at the gearbox output are transmitted to the motor and can be detected by measuring motor currents, where the motor can effectively act as a torque sensor.

\textbf{Actuator Selection.}
We use off-the-shelf mini-cheetah style \cite{Katz2018ALC} brushless actuators with integrated planetary gearboxes Quasi Direct Drive (QDD) actuator as our robot joint modules. These QDDs typically have gear reduction ratios smaller than 10. With high torque brushless motors, these actuators provide sufficient torque for robotic applications while maintaining backdrivability. Although planetary gearboxes have backlash compared to gearboxes such as harmonic drives and cycloid drives, which may reduce precision, recent work utilizing end-to-end imitation learning with tight feedback loops suggests that low precision hardware can still be used for fine-grained tasks with feedback control \cite{myers2023goal,fang2023generalization}. Furthermore, with the rise of legged robots applications, there are now widely available, affordable, and high-performing QDD options on the market. We are using 20Nm gear ratio of 9 actuators for the shoulder joints and 12Nm gear ratio of 40 actuators for the upper arm rotation and elbow joint and 3Nm gear ratio 10 actuators for the lower arm joints as well as the gripper joints, giving an approximate continuous payload of 1kg. The total cost of all 8 QDDs on each arm cost less than \$2000 for our design. For a more detailed cost breakdown, see Table \ref{tab:robot_cost_breakdown}.

\begin{table}[h]
  \small
  \caption{Robot cost breakdown per arm. Our 7 degrees of freedom arm consists of two 20Nm gear ratio 9:1 QDD(quasi-direct-drive) motors as the two shoulder actuators, two 12Nm gear ratio 40:1 QDD as the upper arm rotate and the elbow actuators and three 3Nm gear ratio 10:1 QDD for the lower arm rotate and the wrist yaw and pitch joints. }
  \label{tab:robot_cost_breakdown}
  \centering
  \begin{tabular}{l l l l l c}
    \toprule
    Part Name & QDD 20nm 9:1 & QDD 12nm 40:1 &QDD 3nm 10:1 & Connectors & \textbf{Total Price} \\
    \midrule
    Price Each & \$320 & \$126 & \$95 & \$634 & \\
    Quantity & 2 & 2 & 3 & 1 set & \\
    Price & \$640 & \$252 & \$285 & \$634 & \textbf{\$1811} \\
    \bottomrule
  \end{tabular}
\end{table}

\input{figures/cad_arm/figure}

\textbf{Arm Design.}
\autoref{fig:cad} Shows the dimensions of the robot and the teleoperation device. The specs of our robot and teleoperation device are shown in \autoref{tab:robot_specs}. The bimanual system consists of replicating the robot arm and teleoperation device. We mount them at a 45 degree angle so that the first 2 joints can share the load of gravity. Following \citep{wu2023gello}, we build a scaled kinematic replica of the arm as a teleoperation device and use the same motors we used for the robot arm for the teleoperation device. We set the scale to be 70\% of the original size for easier human teleoperation. This however results in a design constraint, as the distance between joint 6 and joint 7 is 60mm, and with a motor diameter of 57mm, there leaves almost no room to scale down the teaching device. For this DOF we did not scale the kinematics, and found that despite not matching exactly, still resulted in a very capable and intuitive device.

\begin{table*}[h]
  \small
  \caption{Various specifications about the robot and teleoperation device.}
  \label{tab:robot_specs}
  \centering
  \begin{tabular}{l l}
    \toprule
    Robot &  \\
    \midrule
    \midrule
    Mass                   & 4.6kg \\
    Reach (Full Range)     & 572mm \\
    Upper Arm Length       & 279mm \\
    Lower Arm Length       & 252mm \\
    Payload continuous     & $\sim$ 1.5kg  \\
    Payload peak           & $\sim$ 3.5kg  \\
    Degrees of Freedom     & 7     \\
    \midrule
    Teleoperation Device &  \\
    \midrule
    \midrule
    Mass                   & 2.6kg \\
    Upper Arm Length       & 195mm \\
    Lower Arm Length       & 191mm \\
    Degrees of Freedom     & 7 \\
    \bottomrule
  \end{tabular}
\end{table*}

\textbf{Gripper Design.}
Our high-speed gripper features an offset slider-crank linkage design.
This design was chosen for simplicity and robustness over other low cost hands which have more complex mechanisms such as those presented by \citet{guo2019blue, yalehand, shaw2023leaphand}. The finger is designed to grasp on everyday sized objects.
We implement layer jamming, leveraging 3D the printer slicer to generate Finray like geometry from our simple finger shape \cite{khaled@layer_jamming, crooks@finray, roboticfingers}. This design allows for efficient and reliable grasping of various objects. We leverage the same QDD actuators as our arm, which have a nominal RPM of 120, enabling us to achieve a continuous max gripping force of 25 N while being able to close and open under 1 second.

\textbf{Mobility.}
A fully self contained mobile robot solution is important for ease of use and deployment in real world situations as well as the massive expansion of the robots workspace. Oftentimes, mobility in manipulation is achieved through using a mobile robot base, and mounting the rest of the robot system on top of it \cite{bob_mobile, wu2023tidybot, Wyrobek2008TowardsAP}.
In line with the omni-directional base setup described in \cite{xiong2024adaptive}, we integrated our system with the off the shelf AgileX Ranger Mini 2 base, facilitating mobile bimanual manipulation. To expand the robot's vertical operational space and enable the robot to pick up objects on the ground, the bimanual robot is mounted on a gantry, enabling vertical movement. The robot includes an onboard computer and router for Ethernet connectivity without Internet access. The entire mobile platform is powered by the Ranger Mini 2's internal battery and an onboard 24v battery to make it truly wireless.

%% file: figures/appendix_sim_tasks/figure.tex
\begin{figure}[h]
  \centering
  \includegraphics[width=0.83\textwidth]{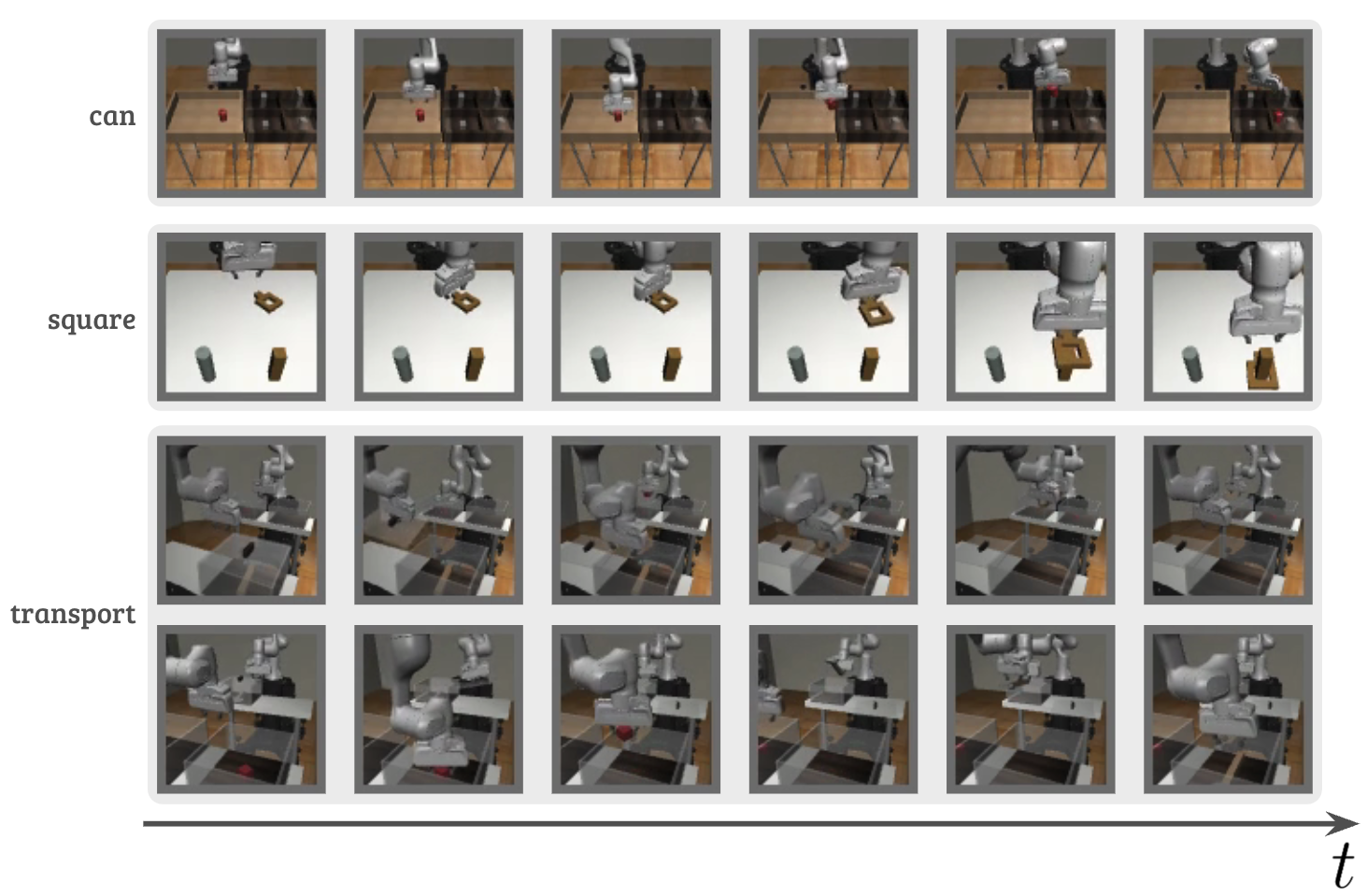}
  \caption{  
  An illustration of the three manipulation tasks we use in Robomimic \citep{mandlekar2021matters}: Pick up a can and place it in a bin; insert a square object into a pole; coordinate two arms to transport a tool.}
  \label{fig:cad}
\end{figure}

%% file: figures/cad_arm/figure.tex
\begin{figure}[h]
  \centering
  \includegraphics[width=0.7\textwidth]{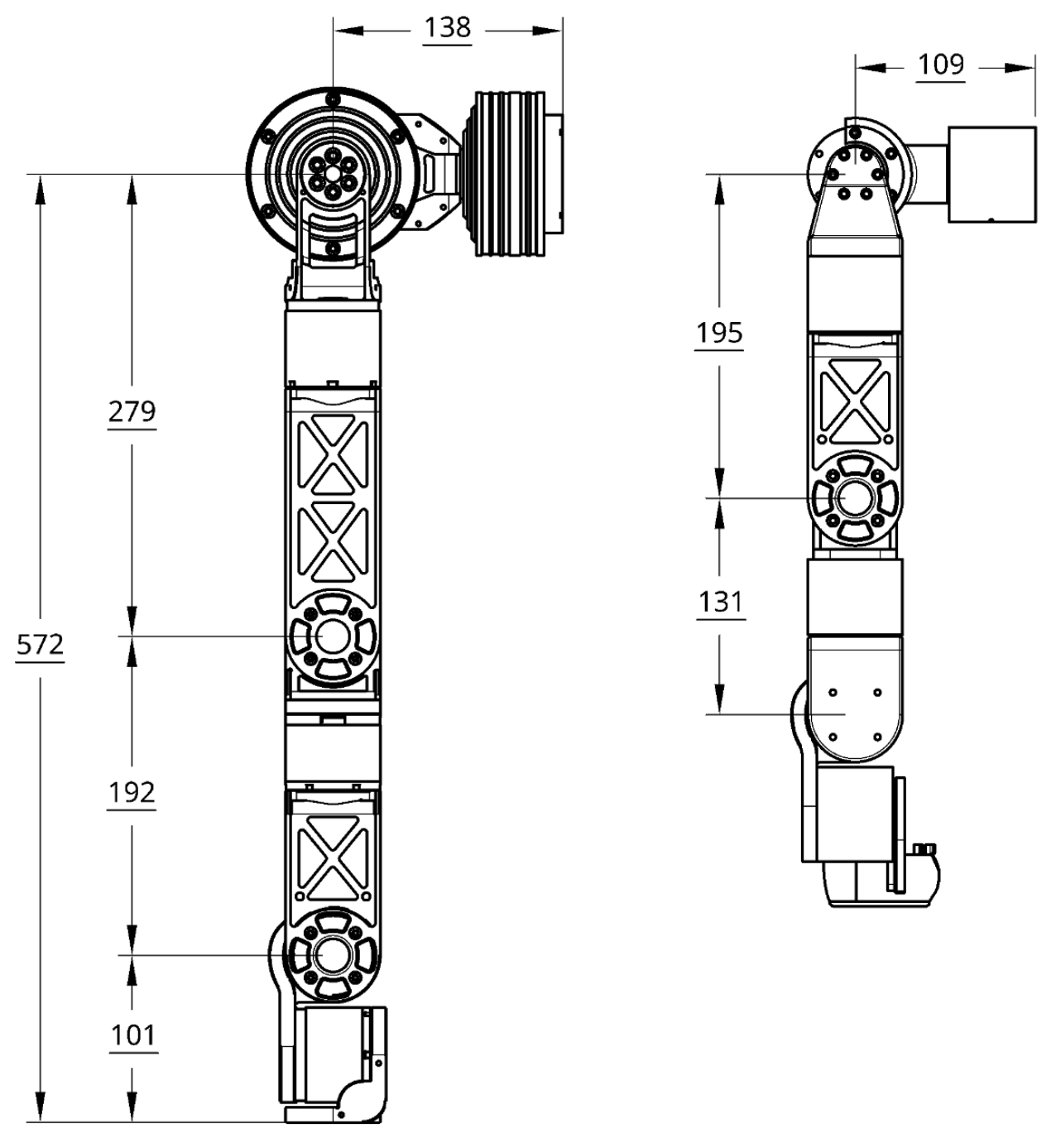}
  \caption{A to scale drawing of our robot arm (left) and teleoperation device (right), with dimensions shown in millimeters.}
  \label{fig:cad}
\end{figure}